# Gradient Similarity: An Explainable Approach to Detect Adversarial Attacks against Deep Learning

Jasjeet Dhaliwal
Center for Advanced Machine Learning
Symantec Corporation
Mountain View, California
Jasjeet_Dhaliwal@symantec.com

Saurabh Shintre
Symantec Research Labs
Symantec Corporation
Mountain View, California
Saurabh_Shintre@symantec.com

## ABSTRACT

Deep neural networks are susceptible to small-but-specific adversarial perturbations capable of deceiving the network. This vulnerability can lead to potentially harmful consequences in security-critical applications. To address this vulnerability, we propose a novel metric called *Gradient Similarity* that allows us to capture the influence of training data on test inputs. We show that *Gradient Similarity* behaves differently for normal and adversarial inputs, and enables us to detect a variety of adversarial attacks with a near perfect ROC-AUC of 95-100%. Even white-box adversaries equipped with perfect knowledge of the system cannot bypass our detector easily. On the MNIST dataset, white-box attacks are either detected with a high ROC-AUC of 87-96%, or require very high distortion to bypass our detector.

## 1 INTRODUCTION

Deep neural networks (DNNs) are known to achieve par-human performance over a wide array of classification tasks such as image recognition and machine translation [24, 41]. However, despite their recent successes, DNNs remain black-box systems due to a limited theoretical understanding of their behavior. In general, a lack of theoretical understanding of a system can be hazardous, as the system may display unpredictable and potentially dangerous behavior on atypical inputs. In the case of DNNs, the unpredictability of their behavior enables an adversary to fool them into misclassification by adding small-but-specific perturbations to regular inputs [9, 16, 35, 42]. The existence of this vulnerability raises critical questions about the reliability, robustness, and safety of DNNs, especially in security-sensitive applications such as autonomous vehicle perception, biometric authentication, medical diagnosis, and malware detection.

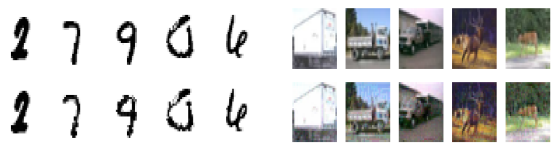

**Figure 1: Original inputs are shown in the first row and the corresponding adversarial inputs in the second row. Each adversarial input leads to an incorrect network classification.**

Adversarial attacks against DNNs are particularly dangerous as they lead to an incorrect prediction with very high network confidence [16, 42] and the perturbations added to the inputs are often imperceptible to human beings, as can be seen in Fig. 1. These attacks can be targeted, i.e., the input can be misclassified into any chosen output class, and also work in a black-box setting where the adversary does not have access to the network's parameters [34, 35]. The existence of such adversarial inputs has also been extended into the physical realm, where adversarial inputs remain adversarial even after physical processing [4, 25, 38]. Given the increasing use of DNNs in real-world applications, it is crucial to design methods capable of detecting and preventing such attacks in order to ensure safe and reliable deployment of DNNs.

To date there are no satisfactory defenses against adversarial attacks on deep learning. Methods that rely on statistical properties of adversarial inputs, e.g., manifold distance and principal components [13, 29], are bypassed by an adversary that has perfect knowledge of the system. Adversarial training [16], the method of creating adversarial inputs and using them to train the network, mainly protects against attack methods considered during training. Robust optimization-based approaches that create provably-secure networks are currently limited to small networks and provide guarantees that are insufficient for practical use [27, 36].

In this work, we draw from the theory of influence functions to detect adversarial inputs. Influence functions are tools from robust statistics that measure the effect of a training point on a test prediction and have recently been used to bring explainability to predictions made by DNNs [22]. They have been used to ascribe predictions back to training data, remove noise from data labels, and can even be used to poison training data to decrease network performance. Given the success of influence functions in explaining regular predictions through the lens of training data, we seek to shed light on the relationship between training data and adversarial predictions using influence. Specifically, we ask the following question:

### Do training points influence predictions on adversarial inputs differently than on normal inputs?

We answer this question in the affirmative and demonstrate that the influence of training points can be used to distinguish adversarial inputs from normal inputs, albeit at a high computational overhead. To alleviate this overhead, we propose a new metric called *Gradient Similarity* (GS) that captures influence at a lower computational cost. Using this metric, we make the following contributions:

- We show that GS allows us to detect adversarial inputs with a **ROC-AUC close to 100%**. Through cross-validation experiments, we demonstrate the success of our detector against unseen attacks.

- We show that our detector is resilient to **white-box attacks**. On the MNIST dataset [26], white-box attacks are either detected with a **ROC-AUC of 87-96%** or require substantial adversarial distortion to bypass our detector.

The paper is organized as follows: we provide a brief survey of existing adversarial attacks and defenses in Section 2. In Section 3, we specify the threat model and notation used in the paper. Section 4 introduces our new metric *Gradient Similarity*, its connections with the theory of influence functions, and an explanation of how we use *Gradient Similarity* to detect adversarial inputs. After describing our experimental setup in Section 5, we evaluate our detector against *zero-knowledge* and *perfect-knowledge* adversaries in Section 6 and 7 respectively. We conclude and discuss directions for future work in Section 8.

## 2 RELATED WORK

We provide a brief overview of existing adversarial attack methods and defenses in this section. A more comprehensive explanation of the attacks is deferred to Section 5.3.

### 2.1 Adversarial attacks

DNNs are vulnerable to small-but-specific perturbations [16, 42] which when added to the input, cause them to misclassify at a very high rate. The success of these attacks detracts from the high accuracy DNNs achieve in a wide variety of tasks, and from their resilience to random noise [12]. A number of efficient attack algorithms have been developed to create such adversarial inputs [9, 16, 25, 30, 31, 34, 35]. Goodfellow *et al.* developed the *Fast Gradient Sign Method* (FGSM) that perturbs the input in the direction that maximizes loss in a single step [16]. Kurakin *et al.* developed *Basic Iterative Method* (BIM) that enhanced the FGSM attack by performing it iteratively, and taking a small step in the direction maximizing loss at each iteration [25]. While these attacks lead to untargeted mis-classification, Papernot *et al.* introduced the Jacobian Saliency Map Attack (JSMA) to mis-classify an input to a pre-specified target class [35]. Further, Papernot *et al.* showed that such attacks can also be launched when the attacker does not have access to the internal parameters of the victim network [34]. As opposed to gradient-based attacks that use a linear approximation of the loss surface, Carlini and Wagner (C&W) utilized optimization to generate adversarial inputs with minimal perturbation [9]. Moosavi-Dezfooli *et al.* created the *DeepFool* (DF) attack that uses a linear approximation of the decision boundary to find adversarial perturbations even when other attacks fail [31]. Such attacks have been demonstrated against neural networks designed for tasks like image and object recognition [16, 42], speech recognition [6], and malware classification [18].

Adversarial attacks can also be successfully launched in the physical world even though the attacker has less control over the input features. For example, Kurakin *et al.* showed that adversarial perturbations remain adversarial even when they are processed by physical media [25]. Sharif *et al.* [38] designed adversarial eye-frames that an attacker can wear to fool a facial recognition system in a live setting. Eykholt *et al.* were able to create adversarial road signs that were misclassified by the computer vision system present in a moving car [11]. Carlini *et al.* demonstrated the success of adversarial attacks against voice assistants, e.g. Google Now, by adding inaudible perturbations to voice commands [6]. Athalye *et al.* generated a 3-D printed adversarial object which fooled a classifier from a wide-range of view points and angles [4]. Grosse *et al.* crafted adversarial inputs to fool neural networks built to classify malware [18]. The success of these attacks in fooling DNNs presents a major roadblock to their safe and reliable deployment.

Due to the adverse ramifications of such attacks and their implications on the safety of deep learning systems, a number of defenses have been proposed in the literature.

### 2.2 Defenses against adversarial attacks

Defenses can generally be divided into two categories: a) defenses that detect adversarial inputs and b) defenses that train networks that are robust to adversarial inputs.

Several adversarial detectors have been proposed in the literature that rely on differences in statistical properties of adversarial and normal inputs. Hendrycks and Gimpel [21] discovered that adversarial inputs have abnormally large magnitudes of low-ranked principal components and used that to separate them from normal inputs. Bhagoji *et al.* used similar data transformations to prevent adversarial attacks [1]. Xu *et al.* used "feature-squeezing", a method of reducing the range of values an input can have, to increase the distortion required for adversarial attacks [44]. Feinman *et al.* used Bayesian uncertainty estimates and the density of hidden layer representations to separate normal and adversarial inputs [13]. Papernot and McDaniel used a *k-nearest neighbors* approach to detect adversarial samples [33]. Metzen *et al.* and Grosse *et al.* trained a classifier to detect adversarial inputs [17, 29]. Meng and Chen proposed a two-pronged defense that first used manifold representations to differentiate normal inputs from adversarial inputs [28]. Then, they created a secondary network to push the adversarial inputs back to the manifold of the normal data. Most of these detectors however rely on statistical properties that are not clearly explainable and may vary from one dataset to another. Further, it was shown that the above detectors can be bypassed by an adversary that can adapt to the defense [7, 8]. He *et al.* further showed that combining a number of weak statistical detectors together also does not lead to a strong detector as an attacker can successfully break an ensemble detector [19].

A different approach to defending against adversarial attacks is to build networks that are robust to such attacks. One technique to do so is adversarial training [16] in which the network designer generates adversarial inputs and uses them during training with correct labels. Adversarial training succeeds in thwarting attacks that were considered during training but is ineffective on other attacks [43]. Tramer *et al.* extended adversarial training by generating adversarial perturbations on other networks and incorporating those perturbations in the training of the original network [43]. Papernot *et al.* proposed to use distillation as a defense [34] which uses a secondary network trained on soft probability labels. However, this approach was found to be ineffective against optimization-based attacks [9]. Recently, robust optimization has been incorporated in network training to build classifiers that are provably resilient to adversarial attacks [20, 27, 36]. Madry *et al.* [27] trained networks with an additional constraint of minimizing the maximum loss in the neighborhood of an input which allowed them to train networks that are robust against adversaries that rely on a first-order approximation of the loss surface. Raghunathan *et al.* [36] developed a differentiable upper-bound on the worst-case loss and were able to minimize it during training to design certifiably robust networks. While robust optimization-based approaches show that training provably-secure neural networks is possible, currently these approaches are limited to small networks and their guarantees are not sufficient for practical use.

We build a detector based on the concept of influence functions. Influence functions try to explain network predictions via the lens of training points [22]. Our detector maintains the explainability of influence functions, achieves high detection rates on numerous adversarial attacks, and is resilient to adaptive attackers. In the next section, we specify the threat network considered in this work and the notation used throughout the paper.

## 3 PRELIMINARIES

### 3.1 Threat model

**Goal of the attacker:** Given a normal input, the goal of the attacker is to create a perturbation, which when added to the input, causes misclassification. For an input x, belonging to class $i$ such that $i \in \{1, \cdots, K\}$, where $K$ is the number of classes, misclassification can be

- **Untargeted:** where the adversarial input can be classified to any class $j$ such that $j \in \{1, \cdots, K\}$ and $j \neq i$
- **Targeted:** where the adversarial input must be misclassified to a pre-specified class $j$ such that $j \in \{1, \cdots, K\}$ and $j \neq i$

**Capabilities of the attacker:** Adversarial capabilities are determined by two factors: the amount of control the attacker has over the input and the amount of knowledge the attacker has about the network and defense. In all cases, we assume that the attacker has complete control of the network input; i.e., the attacker can change any input feature. However, the attacker is constrained by the maximum amount of perturbation that he can add to the input to remain undetected. The adversarial perturbation budget is measured in terms of the maximum permissible norm ($L_1$, $L_2$, or $L_\infty$) of the perturbation. The amount of knowledge the adversary possesses about the network's internal parameters, architecture, and defense leads to three different adversarial scenarios [5, 7].

- *Zero-knowledge adversary:* A zero-knowledge adversary has access to the network's internal parameters and architecture, but has no knowledge that a defense is in place. We refer to attacks launched by such an adversary as "grey-box" attacks.
- *Perfect-knowledge adversary:* A perfect-knowledge adversary has complete knowledge of the network's parameters, architecture, and the defense. This adversary can therefore adapt his attack strategy to account for the defense. This is the strongest possible threat model. Attacks crafted by such an adversary are known as "white-box" attacks.
- *Limited-knowledge adversary:* A limited-knowledge adversary knows that the network is being defended with a given scheme. However, the adversary does not have any knowledge of the network's parameters, training data, or the defense's parameters. Attacks launched by such an adversary are called "black-box" attacks.

In this paper, we do not consider "black-box" attacks as they are the most difficult for an adversary [7].

### 3.2 Notation

We use the following standard notation throughout the paper.

- $C = \{1, 2, \ldots, K\}$ is the set of classes.
- $\mathcal{F}_\theta(\cdot)$ represents the network, where $\theta$ are the parameters of the network.
- $\mathbf{x} \in \mathbb{R}^n$ represents the network input. $\hat{\mathbf{y}} = \mathcal{F}_\theta(\mathbf{x})$, where $\hat{\mathbf{y}} \in [0,1]^K$ is the network output representing a probability vector over $K$ classes. The predicted class for an input $\mathbf{x}$ is computed as arg max($\hat{\mathbf{y}}$).
- $\mathcal{L}(\theta, \mathbf{x}, \mathbf{y})$ represents the loss function for $\mathcal{F}_\theta$ on input $\mathbf{x}$ with respect to the label $\mathbf{y}$. We use categorical cross-entropy as the loss function unless specified otherwise.
- $\mathbb{X}_{tr}$ and $\mathbb{Y}_{tr}$ represent the set of training points and training labels, respectively.
- The symbol $\nabla$ represents the Jacobian; i.e. the first derivative and $\nabla^2$ represents the Hessian; i.e. the second derivative. The subscript of these symbols represents the variable with respect to which the derivative is computed.
- clip($\cdot$) represents the clipping function. For $z \in \mathcal{R}$:

$$\text{clip}(z) = \begin{cases} \text{MIN} & \text{if } z < \text{MIN} \\ z & \text{if MIN} \leq z \leq \text{MAX} \\ \text{MAX} & \text{if } z > \text{MAX} \end{cases}$$

The permissible range of $z$ is [MIN, MAX] and is defined a priori. The clip function is applied element-wise to a vector.

- sign($\cdot$) represents the sign function. For $z \in \mathcal{R}$:

$$\text{sign}(z) = \begin{cases} -1 & \text{if } z < 0 \\ 0 & \text{if } z = 0 \\ 1 & \text{if } z > 0 \end{cases}$$

The function is applied element-wise to a vector.

- We use subscript index notation to index a vector. For instance $\mathbf{x}_{[i]}$ indicates the $i^{\text{th}}$ element of the vector $\mathbf{x}$.

## 4 THEORETICAL FOUNDATIONS OF INFLUENCE AND SIMILARITY

The theoretical understanding of neural networks is currently limited leads to questions about the explainability of decisions made by them. Recently, Koh and Liang used the concept of influence functions to explain the test-time predictions made by a neural network through the lens of its training data [22].

### 4.1 Influence Functions

Influence functions is a tool from robust statistics that provide a way to measure the effect of each training point on the network's test time behavior [10]. Specifically, they measure the change in network loss at a test input when the weight of a specific training point is changed during training. Koh and Liang [22] formalized this concept for DNNs and provided a closed-form expression to measure the influence of a training point on the network loss at a test input. We provide a brief overview of their key results.

Consider the network risk,

$$\mathcal{R}(\theta) = \frac{1}{|\mathbb{X}_{tr}|} \sum_{\substack{\mathbf{x} \in \mathbb{X}_{tr} \\ \mathbf{y} \in \mathbb{Y}_{tr}}} \mathcal{L}(\theta, \mathbf{x}, \mathbf{y})$$

The goal of network training is to find parameters $\theta^*$ that minimize the network risk. Therefore,

$$\theta^* = \arg\min_{\theta} \mathcal{R}(\theta)$$

Assuming that $\mathcal{R}(\theta)$ is convex and differentiable, we have,

$$\nabla_\theta \mathcal{R}(\theta^*) = 0$$

Increasing the weight of a specific training point $\mathbf{x}^*$ by a small amount $\epsilon \in \mathbb{R}$ leads to a new risk function

$$\mathcal{R}_{\mathbf{x}^*, \epsilon}(\theta) = \mathcal{R}(\theta) + \epsilon \mathcal{L}(\theta, \mathbf{x}^*, \mathbf{y}^*)$$

Note that setting $\epsilon = -\frac{1}{|\mathbb{X}_{tr}|}$ is equivalent to leaving the training point $\mathbf{x}^*$ out of the training data. The above formulation leads to a different set of optimal parameters

$$\theta^*_{\mathbf{x}^*, \epsilon} = \arg\min_\theta \mathcal{R}_{\mathbf{x}^*, \epsilon}(\theta)$$

Koh and Liang were able to measure the change in optimal network parameters due to up-weighting $\mathbf{x}^*$ by an infinitesimally small $\epsilon$ as

$$\frac{\partial}{\partial \epsilon} \theta^*_{x, \epsilon} = -H_{\theta^*}^{-1} \cdot \nabla_\theta \mathcal{L}(\theta^*, \mathbf{x}^*, \mathbf{y}^*)$$

where $H_{\theta^*} = \nabla^2_\theta \mathcal{R}(\theta)|_{\theta=\theta^*}$ represents the Hessian matrix of the network risk with respect to the network parameters [22]. Koh and Liang defined the influence of a training point, $\mathbf{x}^*$, on the loss at a test input, $\mathbf{x}'$ as

$$\begin{aligned}\mathcal{I}(\mathbf{x}^*, \mathbf{x}') &\stackrel{\text{def}}{=} \frac{\partial}{\partial \epsilon} \mathcal{L}(\theta^*, \mathbf{x}', \mathbf{y}') \Big|_{\epsilon=0} \\ &= -\nabla_\theta \mathcal{L}(\theta^*, \mathbf{x}', \mathbf{y}')^T \cdot H_{\theta^*}^{-1} \cdot \nabla_\theta \mathcal{L}(\theta^*, \mathbf{x}^*, \mathbf{y}^*)\end{aligned}$$

Thus, the quantity $-\frac{1}{|\mathbb{X}_{tr}|} \mathcal{I}(\mathbf{x}^*, \mathbf{x}')$ measures the change in network loss at $\mathbf{x}'$ when the training point $\mathbf{x}^*$ has been left out from training.

### 4.2 Gradient Similarity

While $-\frac{1}{|\mathbb{X}_{tr}|} \mathcal{I}(\mathbf{x}^*, \mathbf{x}')$ provides a precise analytical formulation to measure the influence of removing a training point, on the loss at a test input, it is a computationally expensive metric. In order to address this issue, we define a new metric that captures the alignment of loss surfaces at training point $\mathbf{x}^*$, and a test input $\mathbf{x}'$, called *Gradient Similarity* as:

$$GS(\mathbf{x}^*, \mathbf{x}') \stackrel{\text{def}}{=} \nabla_\theta \mathcal{L}(\theta^*, \mathbf{x}', \mathbf{y}')^T \cdot \nabla_\theta \mathcal{L}(\theta^*, \mathbf{x}^*, \mathbf{y}^*)$$

Since $\frac{-1}{|\mathbb{X}_{tr}|}$ is a constant, we note that the main difference between influence and gradient similarity is that gradient similarity replaces the vector $H_{\theta^*}^{-1} \cdot \nabla_\theta \mathcal{L}(\theta^*, \mathbf{x}^*, \mathbf{y}^*)$ with the vector $\nabla_\theta \mathcal{L}(\theta^*, \mathbf{x}^*, \mathbf{y}^*)$. Hence, a scaling relationship between these metrics is possible if

$$H_{\theta^*}^{-1} \cdot \nabla_\theta \mathcal{L}(\theta^*, \mathbf{x}^*, \mathbf{y}^*) = \lambda . \nabla_\theta \mathcal{L}(\theta^*, \mathbf{x}^*, \mathbf{y}^*)$$

for a constant $\lambda$. To show equivalence between influence and gradient similarity, we only need to show that the vectors $H_{\theta^*}^{-1} \cdot \nabla_\theta \mathcal{L}(\theta^*, \mathbf{x}^*, \mathbf{y}^*)$ and $\nabla_\theta \mathcal{L}(\theta^*, \mathbf{x}^*, \mathbf{y}^*)$ are scaled versions of one another. That is, for a fixed set of training points

(1) the ratio of norms of the two vectors is constant.
(2) the cosine of the angle between the two vectors is 1.

In order to verify these two properties experimentally, we randomly select a set of 1000 training points from the MNIST dataset [26].

For (1), we compute the ratio $\frac{||\nabla_\theta \mathcal{L}(\theta^*, \mathbf{x}^*, \mathbf{y}^*)||_2}{||H_{\theta^*}^{-1} \cdot \nabla_\theta \mathcal{L}(\theta^*, \mathbf{x}^*, \mathbf{y}^*)||_2}$ across the training points. In our experiments, the ratio was found to be in the range [99.04225, 107.24023], with mean: 100.20286, median: 99.98809 and variance $\approx 0$. We illustrate this effect in Fig. 2 as well. The experiment validates that the ratio of the norms of the two vectors is close to being constant.

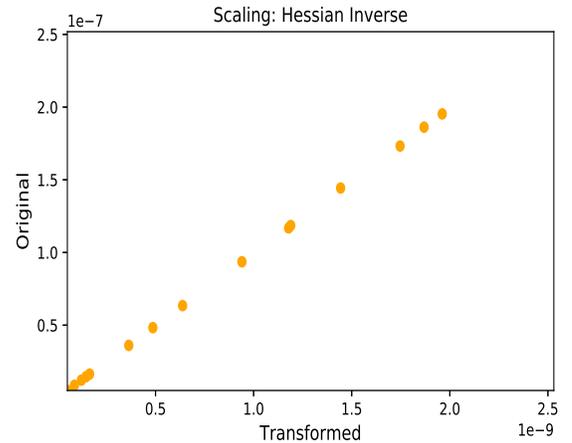

**Figure 2: Norms of $\nabla_\theta \mathcal{L}(\theta^*, \mathbf{x}^*, \mathbf{y}^*)$ (Original) and $H_{\theta^*}^{-1} \cdot \nabla_\theta \mathcal{L}(\theta^*, \mathbf{x}^*, \mathbf{y}^*)$ (Transformed). The plot illustrates the linear relationship between the two vector norms.**

For (2) we compute the cosine-similarity between vectors $H_{\theta^*}^{-1} \cdot \nabla_\theta \mathcal{L}(\theta^*, \mathbf{x}^*, \mathbf{y}^*)$ and $\nabla_\theta \mathcal{L}(\theta^*, \mathbf{x}^*, \mathbf{y}^*)$. We find that the cosine-similarity is in the range [0.99938, 1.0], with mean: 0.99982, median: 0.99986 and variance $\approx 0$. These results experimentally validate that the two vectors are parallel and scaled versions of each other. Note that we expect to see some variance in our results as we use an approximation to $H_{\theta^*}^{-1}$ using Hessian Vector Products (HVP) [22].

These two observations corroborate the equivalence of gradient similarity and influence in measuring relative geometries of the loss surface at a test input and a training point. This equivalence also allows gradient similarity to maintain the explainability of the influence function metric. [1]

---

[1] The role of the $H_{\theta^*}^{-1}$ in scaling the gradient vectors also raises theoretical questions about neural network training. For a matrix $A$, vector $\mathbf{x}$, and a fixed constant $\lambda$, $A\mathbf{x} = \lambda \mathbf{x}$ if and only if $\lambda$ is an eigenvalue of $A$ and $\mathbf{x}$ is an eigenvector of $A$. For an eigenvalue $\lambda$, set of all vectors $\mathbf{x}$ such that $A\mathbf{x} = \lambda \mathbf{x}$ form an eigenspace. Our observations about $H_{\theta^*}^{-1}$ and $\nabla_\theta \mathcal{L}(\theta^*, \mathbf{x}^*, \mathbf{y}^*)$ allude to the possibility that $\nabla_\theta \mathcal{L}(\theta^*, \mathbf{x}^*, \mathbf{y}^*)$ may form

For a fixed training point, gradient similarity can also be further decomposed into two features that are determined solely by the test input: 1) the norm of the gradient vector at the test input and 2) cosine-similarity of the gradient vectors at the test input and the training point. In the section below, we ask if these features can be used to distinguish adversarial and normal inputs. **We answer this question in the affirmative**.

### 4.3 Using Gradient similarity to detect adversarial inputs

Equipped with gradient similarity, which serves as a proxy for influence functions, we ask the following question:

**Does Gradient similarity behave differently for adversarial inputs than for normal inputs ?**

This question essentially asks if there is a fundamental difference between the properties of the loss surface at normal and adversarial inputs. If such a difference exists, then normal inputs would possibly be separable from adversarial inputs. Note that gradient similarity between a test input $\mathbf{x}'$, and training point $\mathbf{x}^*$ can be further decomposed as:

$$GS(\mathbf{x}^*, \mathbf{x}') = ||\nabla_\theta \mathcal{L}(\theta^*, \mathbf{x}', \mathbf{y}')||_2 \, ||\nabla_\theta \mathcal{L}(\theta^*, \mathbf{x}^*, \mathbf{y}^*)||_2 \cos(\alpha_{\mathbf{x}^*, \mathbf{x}'})$$

where, $\alpha_{\mathbf{x}^*, \mathbf{x}'}$ is the angle between the two gradient vectors. In order to measure gradient similarity for a fixed training point with different test inputs, we can ignore $||\nabla_\theta \mathcal{L}(\theta^*, \mathbf{x}^*, \mathbf{y}^*)||_2$ as it is constant. Therefore, the decomposition of gradient similarity allows us to analyze the geometry of the loss surface at $\mathbf{x}'$ using the two features defined below

$$N(\mathbf{x}') \stackrel{\text{def}}{=} ||\nabla_\theta \mathcal{L}(\theta^*, \mathbf{x}', \mathbf{y}')||_2$$
$$C(\mathbf{x}^*, \mathbf{x}') \stackrel{\text{def}}{=} \cos(\alpha_{\mathbf{x}^*, \mathbf{x}'})$$

We experimentally study the behavior of these features for normal and adversarial inputs and show that these features allow us to separate adversarial inputs from normal inputs. First, we describe our experimental setup in the next section.

## 5 EXPERIMENTAL SETUP

### 5.1 Datasets

We consider three datasets for our experiments:

1. **MNIST**: A dataset of 28×28, gray-scale handwritten images containing digits from 0-9 [26]. The dataset contains a total of 70,000 images with ground-truth labels, where the labels can be any digit from 0-9.
2. **CIFAR2**: A dataset of colored images of size $32 \times 32 \times 3$, where the last dimension is for color channels. Since networks that achieve high accuracy on CIFAR10 have a very large parameter space, we adapt our dataset from CIFAR10 [23] in order to meet our computational resource constraints. To create CIFAR2, we randomly picked two classes from CIFAR10 and used the resulting data for our experiments. Hence, CIFAR2 contains data for two classes with 6000 images for each class.
3. **DREBIN**: An Android malware dataset of applications on the Android platform. The dataset contains 123,453 benign applications and 5,560 malicious applications. Each application is represented by a set of binary features collected via malware analysis performed as per [3, 40].

### 5.2 Network setup

We briefly describe the network architecture for each network used in our experiments.

- **MNIST**: We use a Convolutional neural network (CNN) with 2 convolutional layers, and 2 fully connected layers, for a total of 4 layers. At each layer, we use the ReLU activation function. The first convolutional layer uses 32 channels, and the second convolutional layer uses 64 channels. Both convolutional layers use a kernel of size 5 × 5 with no padding. We apply dropout with probability 0.5, after the the second convolutional layer, and the first fully connected layer. In order to turn the scores of the second fully connected layer into probabilities, we apply the standard softmax function. The network achieves an accuracy of 99.30 % on a holdout test set of 6055 samples.
- **CIFAR2**: We use the same network architecture as MNIST with the following changes. Each convolutional layer uses a kernel size of 3×3 with padding added in order to ensure that the output size of the convolutional layer is the same as its input size. The network achieves an accuracy of 98.72 % on a holdout test set of 1218 samples.
- **DREBIN** We use a Fully-Connected network with 4 layers. At each layer we apply the ReLU activation function, followed by dropout with probability 0.5. The network achieves an accuracy of 97.37 % on a holdout test set of 3913 samples.

Our code and trained models have been made publicly available.

### 5.3 Attacks

We consider six different attacks: for the untargeted setting, we consider FGSM [16], two versions of the BIM [25] attack: BIM-A and BIM-B, and the DeepFool attack [31]. In the targeted setting, we consider the JSMA [35] attack and the C&W attack [9]. While these attacks traditionally assume the input features to be continuous, this assumption is violated for DREBIN dataset, in which input features take binary values. Therefore, the attack algorithms have to be modified to work on DREBIN inputs.

**FGSM**: FGSM [16] computes the gradient of the model loss with respect to input features and perturbs each feature in a way that increases model loss. The parameter $\epsilon$ controls the amount of perturbation added to each feature. For an input $\mathbf{x}$ and its true label $\mathbf{y}$, the corresponding adversarial input $\mathbf{x}_{\text{adv}}$ is computed as:

$$\mathbf{x}_{\text{adv}} = \text{clip}(\mathbf{x} + \epsilon.\text{sign}(\nabla_\mathbf{x} \mathcal{L}(\theta^*, \mathbf{x}, \mathbf{y})))$$

For DREBIN, we select all the input features $\mathbf{x}_{[i]}$ for which

$$\text{sign}(\nabla_\mathbf{x} \mathcal{L}(\theta^*, \mathbf{x}, \mathbf{y}))) > 0$$

We then flip those feature values by changing 0 to 1, or 1 to 0.

**BIM-A, BIM-B**: Basic Iterative Method is an iterative version of the FGSM attack such that it applies a smaller perturbation in each iteration [25]. There are two possible versions of this attack: BIM-A,

---

an eigenspace of $H_{\theta^*}^{-1}$ and hence, of its inverse $H_{\theta^*}$, the Hessian of $\mathcal{R}(\theta^*)$. The cause of this phenomenon is unknown to us and requires further exploration.

| Dataset | Original | FGSM | | BIM-A | | BIM-B | | JSMA | | C&W | | DF | |
|---|---|---|---|---|---|---|---|---|---|---|---|---|---|
| | Acc. | $L_2$ | Acc. | $L_2$ | Acc. | $L_2$ | Acc. | $L_2$ | Acc. | $L_2$ | Acc. | $L_2$ | Acc. |
| MNIST | 100.00% | 5.74 | 9.00% | 2.49 | 0.50 % | 4.62 | 0.50 % | 4.69 | 2.10 % | 3.64 | 0.40 % | 1.66 | 0.40 % |
| CIFAR2 | 98.50 % | 5.43 | 0.00 % | 1.32 | 0.00 % | 4.42 | 0.00 % | 3.66 | 1.50 % | 2.01 | 5.00 % | 1.03 | 1.50 % |
| DREBIN | 91.00 % | 615.04 | 50.00 % | 1.00 | 0.00 % | 2.40 | 25.00 % | 1.33 | 0.00 % | – | – % | – | – % |

Table 1: Prediction accuracy of our undefended networks on the original inputs and their adversarial versions. We also report the average $L_2$ distortion of the misclassified inputs.

where the adversary stops the iterations as soon as misclassification is achieved and BIM-B, where the adversary keeps iterating for a fixed number of iterations. For DREBIN, we update BIM-A and BIM-B by selecting the input feature $\mathbf{x}_{[i]}$ at each iteration such that

$$i = \arg\max_i \ (\nabla_\mathbf{x} \mathcal{L}(\theta^*, \mathbf{x}, \mathbf{y})_{[i]})$$

We then flip the feature value from 0 to 1 or 1 to 0.

**JSMA**: While FGSM and BIM attacks lead to untargeted misclassification, Papernot et al. developed an iterative method to achieve misclassification of an input to any pre-specified class. Their attack takes the target class $t$ as input and outputs an adversarial sample $\mathbf{x}_{adv}$ such that $\arg\max(\mathcal{F}_{\theta^*}(\mathbf{x}_{adv})) = t$. In each iteration of the algorithm, a Jacobian-based saliency map $S(\mathbf{x}, t)$ is created and the feature with the max value in this map, is perturbed by $\epsilon$. The saliency map is created as follows:

$$S(\mathbf{x}, t)_{[i]} = \begin{cases} 0, & \text{if } \frac{\partial(\mathcal{F}_{\theta^*}(\mathbf{x})_{[t]})}{\partial x_i} < 0 \text{ or } \sum_{j \neq t} \frac{\partial(\mathcal{F}_{\theta^*}(\mathbf{x})_{[j]})}{\partial x_i} > 0 \\ \left|\frac{\partial(\mathcal{F}_{\theta^*}(\mathbf{x})_{[t]})}{\partial x_i}\right| \left|\sum_{j \neq t} \frac{\partial(\mathcal{F}_{\theta^*}(\mathbf{x})_{[j]})}{\partial x_i}\right|, & \text{otherwise} \end{cases}$$

For DREBIN, we modify JSMA, so as to flip the feature values from 0 to 1, or 1 to 0 in the perturb step. Further, we only modify each input feature at most once.

**C & W**: Carlini and Wagner built an optimization-based attack where the goal is to find the smallest perturbation that can cause a misclassification [9]. The C&W attack works in both targeted and untargeted settings and works with a variety of distance metrics. Here we only consider their $L_2$ attack which can be described as an optimization problem:

$$\min_{\mathbf{x}_{adv}} ||\mathbf{x}_{adv} - \mathbf{x}||_2 + c \cdot g(\mathbf{x}_{adv})$$

where,

$$g(\mathbf{x}_{adv}) = \max(\max\{Z(\mathbf{x}_{adv})_{[i]} : i \neq t\} - Z(\mathbf{x}_{adv})_{[t]}, -\kappa),$$

$Z()$ represent the pre-softmax outputs or *logits* of the network. $\kappa$ represents the minimum acceptable network confidence for the adversarial perturbation, and $c$ is a weighting constant used during optimization.

**DeepFool**: Moosavi-Dezfooli et al. [31] used a linear approximation of the network decision boundary. Their algorithm works in an iterative manner, and at each iteration finds the perturbation that brings the input closest to a linear approximation of the decision boundary. The algorithm terminates once misclassification is achieved.

In our experiments, the following attack parameters were used:

- **MNIST**: We set $\epsilon = 0.3$ for FGSM, BIM-A, and BIM-B. For BIM-A, and BIM-B, we set the maximum iterations to 10. For C&W we set the maximum iterations to 100.
- **CIFAR2**: We use the same attacks and settings as in MNIST, however we set $\epsilon = 0.1$ as a higher $\epsilon$ leads to images that are unrecognizable even by human beings.
- **DREBIN** We test the DREBIN network on the modified versions of FGSM, BIM-A, BIM-B, and JSMA. We use 10 iterations for BIM-A, BIM-B.

We use a modified version of the cleverhans library [32] to implement our attacks and include the modified version in our publicly available source code. Table 1 reports the accuracy of our undefended networks on these attacks and the average $L_2$ distortion added by each attack. Clearly, these attacks succeed in fooling our undefended networks at a high rate. Sample adversarial images created by these attacks for MNIST and CIFAR datasets are displayed in Fig. 3.

## 6 DETECTING GREY-BOX ATTACKS

Zero-knowledge adversaries have access to the networks's parameters and architecture but do not have any knowledge of the defense mechanism. To evaluate the efficacy of our approach against such adversaries, we first analyze the behavior of the $N(\mathbf{x}')$ and $C(\mathbf{x}, \mathbf{x}')$ for normal and adversarial inputs. To do so, we select a random subset of training points and test inputs from each dataset. For each normal test input we create the corresponding adversarial input for each attack method. For all test inputs, the values of $N(\mathbf{x}')$ are visualized in Fig. 4(a). As can be seen, the norm of the gradient vectors is low for all normal points and high for all adversarial points except for BIM-B. We also plot $\max_{\mathbf{w} \in \mathbb{X}_{tr}} C(\mathbf{w}, \mathbf{x}')$ for each of the test inputs in Fig 4(b). The graph shows that the max cosine-similarity values are close to 1 for normal inputs and smaller for adversarial inputs from all attack methods, including BIM-B. One can note from the graph that the norm and max cosine-similarity for gradient vectors at the original test inputs lie in a separate band as compared to their adversarial counterparts.

We also highlight the above phenomenon by evaluating the percentage of test inputs for which the norm ratio: $\frac{N(\mathbf{x}'_{adv})}{N(\mathbf{x}')} > 1$. We do the same for max cosine-similarity and evaluate the percentage of test inputs for which the max cosine-similarity ratio: $\frac{\max_{\mathbf{w} \in \mathbb{X}_{tr}} C(\mathbf{w}, \mathbf{x}'_{adv})}{\max_{\mathbf{w} \in \mathbb{X}_{tr}} C(\mathbf{w}, \mathbf{x}')} < 1$.

As reported in Table 2, for all attack methods except BIM-B, the norm of the gradient vector of an adversarial input is larger than the norm of the gradient vector for the corresponding normal input.

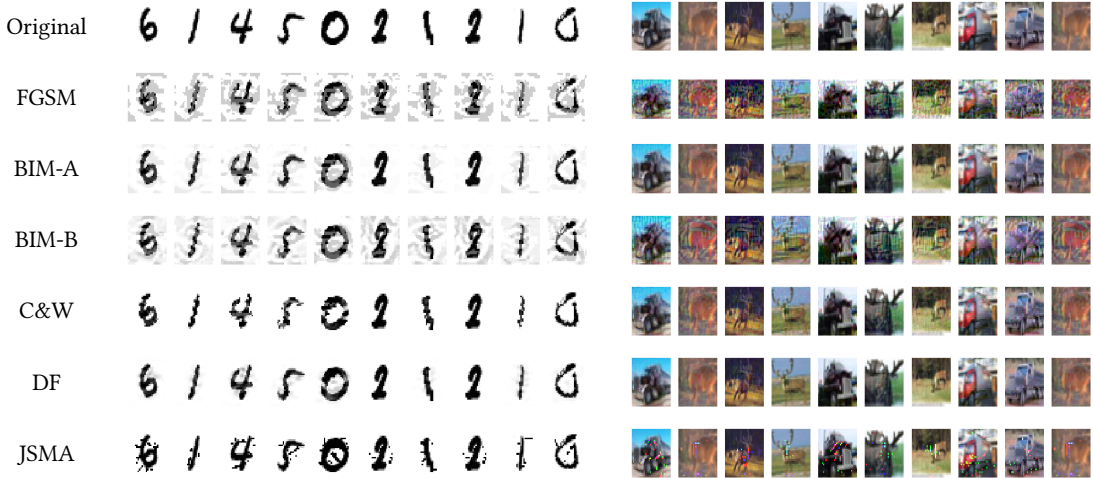

Figure 3: Grey-box adversarial images for MNIST and CIFAR2

This behavior is caused by the fact that adversarial inputs crafted by most attack methods are close to the classification boundary. This in turn means that the loss at those inputs is sensitive to a small change in the network parameters and therefore, the norm of gradient vectors for such inputs is higher. The BIM-B attack pushes inputs farther away from the classification boundary and into regions of the decision surface that are not as sensitive to changes in the classification boundary. Therefore, the gradient vectors of adversarial inputs created by BIM-B display a smaller norm.

| Input | MNIST $\frac{N(x'_{adv})}{N(x')} > 1$ | CIFAR2 $\frac{N(x'_{adv})}{N(x')} > 1$ | DREBIN $\frac{N(x'_{adv})}{N(x')} > 1$ |
|---|---|---|---|
| FGSM | 95 % | 21% | 10 % |
| BIM-A | 98 % | 93% | 44% |
| BIM-B | 64 % | 2% | 50 % |
| JSMA | 99 % | 99 % | 40% |
| C&W | 99 % | 88 % | – |
| DF | 99 % | 100 % | – |

Table 2: Fraction of adversarial inputs having higher gradient norm compared to their normal counterparts. x represents the original input and x′ represents the adversarial version of that input.

| Input | MNIST $\frac{MC(x_{adv})}{MC(x')} < 1$ | CIFAR2 $\frac{MC(x_{adv})}{MC(x')} < 1$ | DREBIN $\frac{MC(x_{adv})}{MC(x')} < 1$ |
|---|---|---|---|
| FGSM | 93 % | 69 % | 100% |
| BIM-A | 97% | 79% | 100% |
| BIM-B | 93 % | 64% | 100 % |
| JSMA | 98 % | 87% | 100% |
| C&W | 98 % | 82 % | – |
| DF | 98 % | 84 % | – |

Table 3: Fraction of adversarial inputs with lower maximum cosine-similarity than the corresponding normal inputs. Here, we use $MC(x') = \max_{w \in \mathbb{X}_{tr}} C(w, x')$, where the maximum is taken over all the training points w. x represents the original input and x′ represents the adversarial version of that input.

In contrast, the maximum cosine-similarity for a normal input is found to be higher than its adversarial counterpart for all attack method as reported in Table 3, including BIM-B. This suggests a lack of alignment between the loss surface at adversarial inputs and the loss surface at training points.

Given the above observations, we build a detector that utilizes cosine-similarity values as features in order to detect adversarial inputs. More specifically, we use a logistic regression detector that outputs a binary label indicating whether an input is adversarial or normal. The input to the detector is a feature vector consisting of the cosine-similarity values for a test input with a fixed subset of the training data. We train the detector on feature vectors created from a mixture of normal inputs and adversarial inputs. The performance of our detector, measured in ROC-AUC, on each grey-box attack is found to be very high, as reported in Table 5. Even when the adversarial class was composed of inputs created using different attack algorithms, the detector was able to detect all adversarial inputs with near-perfect ROC-AUC. For these scenarios, the ROC curves are presented in Fig. 5 and the ROC-AUC values are reported as 1.0 for MNIST, .9966 for CIFAR2, and 1.0 for DREBIN.

In order to test whether our approach is able to detect new adversarial attacks, we perform leave-one-out cross validation. In this scenario, we train the detector on regular inputs and adversarial inputs from five out of the six grey-box attack methods. We then test the detector on normal inputs and on adversarial inputs from the sixth attack method that was left out during training. We report the results of this experiment in Table. 4. As can be seen, our detector

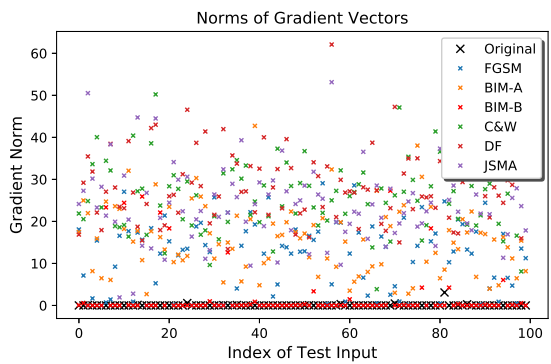

(a) Gradient Norm

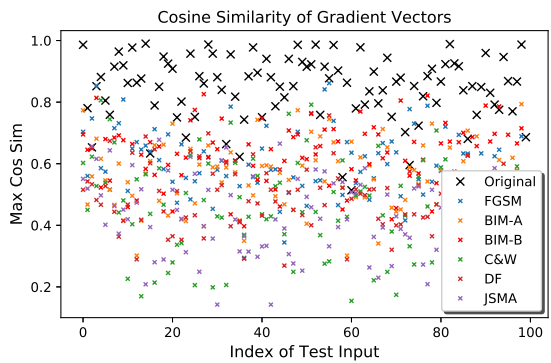

(b) Maximum cosine-similarity

Figure 4: Geometric properties of the network loss surface for MNIST. As can be seen in (a) the norm of gradient vectors for the normal inputs forms a band below the norms for the corresponding adversarial inputs. Similarly in (b) we show that the gradient vectors at the normal inputs have a higher cosine-similarity to training points than their adversarial counterparts.

| Dataset | FGSM Acc. | BIM-A Acc. | BIM-B Acc. | JSMA Acc. | C&W Acc. | DF Acc. |
|---|---|---|---|---|---|---|
| MNIST | 99.27 | 97.22 | 78.06 | 97.69 | 96.65 | 99.90 |
| CIFAR2 | 98.10 | 97.60 | 99.50 | 98.55 | 97.14 | 99.51 |
| DREBIN | 98.18 | 100.00 | 93.75 | 100.00 | – | – |

Table 4: Leave-one-out cross validation: For each dataset and grey-box attack, we report the accuracy (in %) of our detector in detecting the grey-box attack that was left out from the training set of the detector.

shows high accuracy in detecting each attack that was left out from its training set.

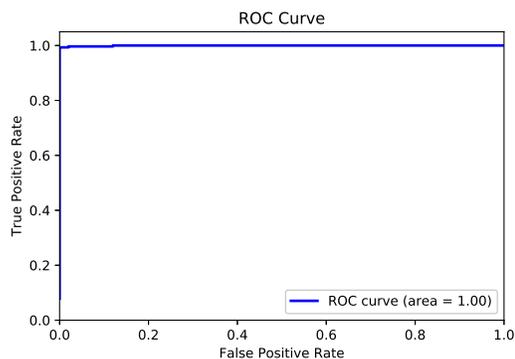

(a) MNIST

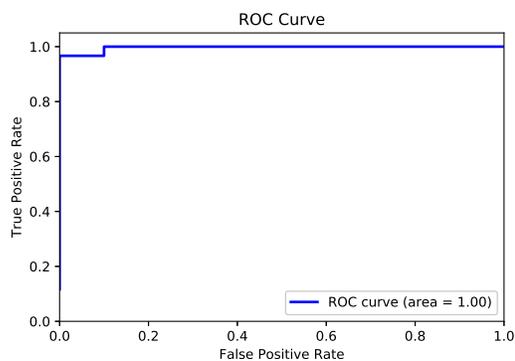

(b) CIFAR2

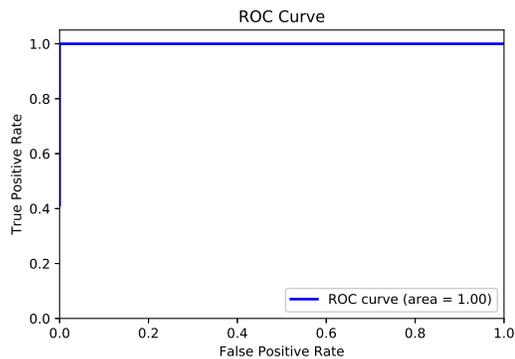

(c) DREBIN

Figure 5: ROC curves for logistic regression grey-box detector

## 7 DETECTING WHITE-BOX ATTACKS

One of the biggest challenges in detecting adversarial inputs is the inability of most detectors to withstand *perfect-knowledge* adversaries [7]. Adversaries equipped with knowledge of the network parameters, architecture, and defense can create adversarial inputs that mimic properties of normal inputs that detectors rely on. It has been argued that a *perfect-knowledge* adversary is an impractical threat model as machine learning systems are typically deployed as black-boxes. However, network parameters and training data can be leaked even in a black-box setting through model inversion and membership inference attacks [14, 39]. While privacy-enhancing technologies, such as differential privacy [2], can prevent such leakage, network parameters can still be leaked by an insider with access to the network. Therefore, it is critical to evaluate the robustness of adversarial detectors against *perfect-knowledge* adversaries. We evaluate the efficacy of our detector under two different types of white-box attacks that can be launched by a *perfect-knowledge* attacker. These attack methods are inspired by white-box attacks developed by Carlini and Wagner [7], Sabour *et al.* [37], and Ghorbani *et al.* [15].

| Dataset | FGSM | | BIM-A | | BIM-B | | JSMA | | C&W | | DF | |
|---|---|---|---|---|---|---|---|---|---|---|---|---|
| | $L_2$ | AUC | $L_2$ | AUC | $L_2$ | AUC | $L_2$ | AUC | $L_2$ | AUC | $L_2$ | AUC |
| MNIST | 5.74 | 99.96% | 2.49 | 100.00% | 4.62 | 99.56% | 4.69 | 100.00% | 3.64 | 100.00% | 1.66 | 100.00% |
| CIFAR2 | 5.43 | 99.00% | 1.32 | 99.00% | 4.42 | 100.00% | 3.66 | 100.00% | 2.01 | 100.00% | 1.03 | 100.00% |
| DREBIN | 615.04 | 100.00 % | 1.00 | 100.00 % | 2.40 | 100.00 % | 1.33 | 100.00 % | – | – % | – | – % |

Table 5: Grey-box Detection: The first sub-column shows the average $L_2$ distortion of an adversarial input and the second sub-column lists the ROC-AUC of the logistic regression detector.

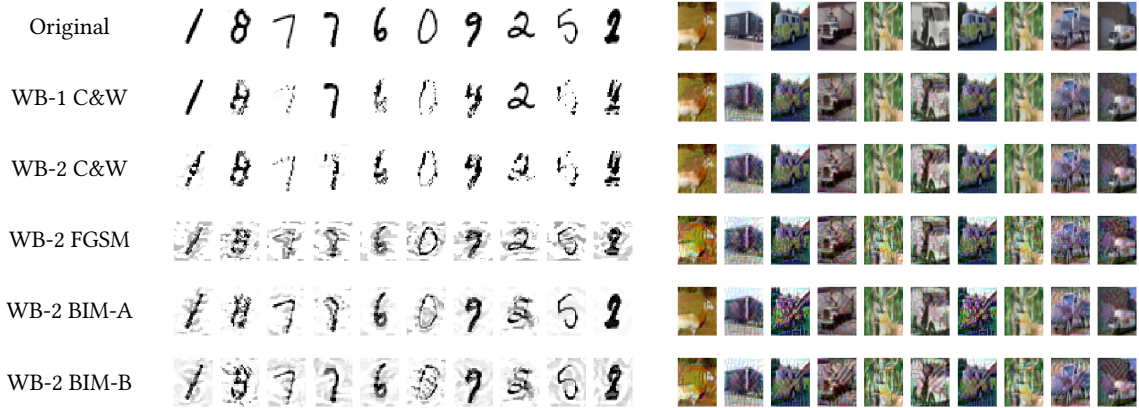

Figure 6: White-box adversarial images for MNIST and CIFAR2

| Dataset | WB-1 C&W | | WB-2 C&W | | WB-2 FGSM | | WB-2 BIM-A | | WB-2 BIM-B | |
|---|---|---|---|---|---|---|---|---|---|---|
| | $L_2$ | AUC | $L_2$ | AUC | $L_2$ | AUC | $L_2$ | AUC | $L_2$ | AUC |
| MNIST | 6.17 | 93.14% | 4.13 | 96.40% | 9.79 | 96.40% | 10.85 | 96.40% | 10.99 | 87.45% |
| CIFAR2 | 5.47 | 56.83% | 4.47 | 56.49% | 5.61 | 50.99% | 5.16 | 59.00% | 4.61 | 50.00% |

Table 6: White-box detection: The first sub-column shows the average $L_2$ distortion of succcessful adversarial inputs and the second sub-column lists the ROC-AUC of the threshold detector.

For the *perfect-knowledge* attacker, we build a detector that uses $\max_{\mathbf{w} \in \mathbb{X}_{tr}} (C(\mathbf{w}, \mathbf{x}'))$ and $N(\mathbf{x}')$ as features for a test input $\mathbf{x}'$, and sets thresholds for both features that separate normal and adversarial inputs. The attacker is aware of the features and thresholds that are being used. With this knowledge, the attacker can create his adversarial inputs either in one phase or two phases. The attack is considered successful only if the adversarial input is mis-classified by the network and is labeled as normal by the detector.

**One-phase attack:** Here the approach of the adversary is to create a single objective function that can be optimized to the cause mis-classification and bypass the detector simultaneously. For our detector, the goal of the attacker is to craft adversarial inputs that also have a small norm of the loss gradient and high cosine-similarity with at least one training point of the target class.

To do so, the attacker must craft an $\mathbf{x}'$ such that $N(\mathbf{x}')$ is minimized and $C(\mathbf{x}_g, \mathbf{x}')$ is maximized with respect to a guide input $\mathbf{x}_g$. The guide input $\mathbf{x}_g$ can be selected by the adversary as the input with which the original input, $\mathbf{x}$, already has the highest cosine-similarity; i.e., $\mathbf{x}_g = \arg\max_{\mathbf{w} \in \mathbb{X}_{tr}}(C(\mathbf{w}, \mathbf{x}))$. These two additional terms can be incorporated in the optimization framework of the C&W attack to create the white-box C&W attack (WB-1 C&W). Let $l(\mathbf{x}')$ represent the objective function to be minimized by a regular C&W attack to cause misclassification. Then, the updated objective function for the WB-1 C&W attack can be formulated as:

$$\tilde{l}(\mathbf{x}') = l(\mathbf{x}') + \mathrm{N}(\mathbf{x}') + \max(-\mathrm{C}(\mathbf{x}_g, \mathbf{x}') + 1.0, 0.0)$$

**Two-phase attack:** In phase one of this attack, the attacker first uses a standard grey-box attack to create an adversarial input that is misclassified. For our analysis, we consider to FGSM, BIM-A, BIM-B, and C&W attacks for phase one. In the second phase of the attack, a C&W attack with a modified objective is used to minimize the norm and maximize the cosine-similarity features while maintaining misclassification. Let $\mathbf{x}'_{(1)}$ be the intermediate adversarial input created in phase one. Then, the modified objective for phase two is:

$$l(\mathbf{x}'_{(1)}) = \mathrm{N}(\mathbf{x}'_{(1)}) + \max(-\mathrm{C}(\mathbf{x}_g, \mathbf{x}'_{(1)}) + 1.0, 0.0)$$

where the constraint of misclassification in the updated loss function is maintained. The selection of the guide input $\mathbf{x}_g$ in two-phase attack differs from one-phase attacks in that $\mathbf{x}_g$ is the training point that has the maximum cosine-similarity with the adversarial input produced in phase one by the grey-box attack. That is,

$$\mathbf{x}_g = \arg\max_{\mathbf{w} \in \mathbb{X}_{tr}} (\ \mathrm{C}(\mathbf{w}, \mathbf{x}'_{(1)})$$

We call our two-phase attacks: WB-2 FGSM, WB-2 BIM-A, WB-2 BIM-B, and WB-2 C&W based on the corresponding grey-box attack used in phase one. The images created by these attacks are presented in Fig. 6. Due to the binary nature of the DREBIN dataset, creating these white-box attacks on the DREBIN dataset requires significant changes to our optimization algorithms, and hence we leave DREBIN out of our present evaluation.

The results of our evaluation on the MNIST and CIFAR2 datasets are presented in the form of ROC-AUC calculations reported in Table 6, along with the $L_2$ distortion of the adversarial inputs that bypass our detector. As can be seen, our method is able to successfully detect all white-box attacks on the MNIST datasets. The ROC-AUC of our detector is found to be 87-97%. For practical white-box attack approaches, these results represent an improvement over guarantees given by robust-optimization based defenses [27, 36].

The results for CIFAR2 are less promising. We detect different attacks with an AUC of 50-59%. The poor performance of our detector on CIFAR2 is not fully explainable and we postulate that it is caused by two reasons: a) the number of training points used for cosine-similarity scores is low due to computational resource constraints and b) the distribution of the CIFAR data has a very large support in the input space. This is also the reason why a number of other detectors fare poorly on CIFAR [7]. We leave further exploration of these observations as future work. Despite the poor performance of our detector on CIFAR images, the success of our detector on the MNIST dataset shows that building statistical detectors that are resilient to white-box attacks is possible.

## 8 CONCLUSIONS AND FUTURE WORK

We proposed a new defense against adversarial attacks on DNNs based on the geometry of the loss surface and the theory of influence functions. We introduced a new metric called *Gradient Similarity* which substitutes the computationally-expensive metric of influence. A logistic-regression detector based on features that emanate from gradient similarity allows us to detect inputs crafted by a *zero-knowledge* adversary with very high ROC-AUC. Moreover, our threshold detector is able to detect white-box attacks on the MNIST dataset with ROC-AUC ranging from 87-97%. Our detector requires little-to-no-change in the primary network which makes it more practical when compared to approaches like robust optimization that require a substantial change in network training.

However, our detector suffers from a few limitations. It is unable to detect white-box attacks on CIFAR2. While cross-validation experiments shows that our detector is able to detect new attack methods, it has not been evaluated against all possible attack strategies and therefore, no formal claims about its robustness can be made. The successes and limitations of our detector also raise a number of open questions around the theory of neural networks. Specifically, the relationship between network predictions and training points that exert maximum influence is poorly understood. The equivalence of behavior between influence and gradient similarity also raises questions about the properties of the Hessian of the risk function and its eigenspaces. Exploring the answers to these questions can lead to a better understanding of neural network predictions and open doors for the development of strong defenses.

## ACKNOWLEDGEMENTS

We would like to thank Symantec Corporation for supporting this research.